\documentclass[conference, compsoc]{IEEEtran}
\ifCLASSOPTIONcompsoc
  \usepackage[nocompress]{cite}
\else
  \usepackage{cite}
\fi
\ifCLASSINFOpdf
  \usepackage[pdftex]{graphicx}
\else
  \usepackage[dvips]{graphicx}
\fi
\usepackage{amsmath}
\usepackage{url}

\usepackage{multirow}
\usepackage{amsfonts}
\usepackage{amssymb}
\usepackage{booktabs}
\usepackage{float}

\begin{document}
%
\title{Contrastive Learning-Based privacy metrics in Tabular Synthetic Datasets}


%

\author{\IEEEauthorblockN{
Milton Nicol\'as Plasencia Palacios\IEEEauthorrefmark{1} \IEEEauthorrefmark{2} \IEEEauthorrefmark{3},
Sebastiano Saccani\IEEEauthorrefmark{1},
Gabriele Sgroi\IEEEauthorrefmark{4}\\
Alexander Boudewijn\IEEEauthorrefmark{1} and
Luca Bortolussi\IEEEauthorrefmark{2}
} \\
\IEEEauthorblockA{
\IEEEauthorrefmark{1} Aindo, Trieste, Italy \\
\IEEEauthorrefmark{2} University of Trieste, Trieste, Italy \\
\IEEEauthorrefmark{4} Work done while at Aindo, \\
\IEEEauthorrefmark{3} Corresponding author, plasencia.milton@gmail.com \\
}
}




\maketitle

\begin{abstract}
Synthetic data has garnered attention as a Privacy Enhancing Technology (PET) in sectors such as healthcare and finance. When using synthetic data in practical applications, it is important to provide protection guarantees. In the literature, two family of approaches are proposed for tabular data: on the one hand, Similarity-based methods aim at finding the level of similarity between training and synthetic data. Indeed, a privacy breach can occur if the generated data is consistently too similar or even identical to the train data. On the other hand, Attack-based methods conduce deliberate attacks on synthetic datasets. The success rates of these attacks reveal how secure the synthetic datasets are. 

 In this paper, we introduce a contrastive method that improves privacy assessment of synthetic datasets by embedding the data in a more representative space. This overcomes obstacles surrounding the multitude of data types and attributes. It also makes the use of intuitive distance metrics possible for similarity measurements and as an attack vector. In a series of experiments with publicly available datasets, we compare the performances of similarity-based and attack-based methods, both with and without use of the contrastive learning-based embeddings. Our results show that relatively efficient, easy to implement privacy metrics can perform equally well as more advanced metrics explicitly modeling conditions for privacy referred to by the GDPR.
\end{abstract}

\IEEEpeerreviewmaketitle

%

\section{Introduction}

The merits of Artificial Intelligence (AI) and advanced data analytics are increasingly gaining widespread recognition. This results in a strong demand for reliable data in important industries such as healthcare and finance. Unfortunately, the required data is often sensitive in nature, limiting its mobility and use. 

An approach to overcome these obstacles that is quickly gaining traction is substituting real data by realistic \emph{synthetic data}~\cite{vanbreugel2023privacy}. Such data is completely artificial, not or minimally pertaining to real individuals. However, it can be constructed so that it retains the patterns required for analysis and AI development with an unassailable degree of realism. Deep-learning-based generative models to do so include Generative Adversarial Networks (GANs, see~\cite{orig_Gan, orig_GAN_tab, orig_CTGAN_tabular}); Variational AutoEncoders (VAE, see~\cite{orig_VAE, orig_b_VAE, PanfIEEE}); transformers~\cite{solatorio2023realtabformer}; and diffusion models~\cite{Diffusion, kotelnikov2022tabddpm}.

When using synthetic data as a Privacy Enhancing Technology (PET), it is important to understand the degree to which it protects real data subjects' right to privacy. Multiple methods for measuring the degree of privacy protection of a given synthetic dataset are based on distance metrics. Among the most prominent are distance to closest record (DCR) metrics and deliberate attacks relying on distances. 

Unfortunately, tabular (synthetic) data often contains many attributes and data types. This leads to numerous obstacles for privacy assessment. Firstly, the plurality of types hinders the use of standardized distance metrics, such as the euclidean distance. Secondly, the large dimensionality may obscure patterns that require semantic understanding of attribute names to identify. Thirdly, the \emph{curse of dimensionality} may lead to unrealistic assessments.

In this paper, we develop a novel contrastive learning method to overcome these obstacles. Our method improves distance-based privacy assessment of synthetic datasets by embedding the data in a more representative space. The method represents a given tabular synthetic dataset in an embedding in which euclidean distances between records represent quantitative and qualitative similarities. existing methods such as the DCR and deliberate attacks can leverage this space to improve their efficacy. After describing the method, we compare the DCR and the attack-based metrics from~\cite{Anonymeter} both with and without using the contrastive learning-based embeddings.

\section{Relation to prior research}

Embedding spaces obtained through contrastive learning are used for outlier detection~\cite{shenkar2021anomaly, qiu2022neural}. The underlying intuition is the same, as embeddings may unveil unique patterns that are difficult to detect in datasets' original representations. Though similarity-based metrics are widespread in synthetic data privacy assessment~\cite{boudewijn2023privacy, OtherSurvey}, the use of embeddings remains underexplored. Guillaudeux et al.~\cite{avatar} used a dimensionality reduction (FAMD) method to evaluate distance-based privacy metrics. However, our contrastive learning-based method is more autonomous and allows for the incorporation of more qualitative information. 

Giomi et al.~\cite{Anonymeter} use an empirical attack framework for singling out, linkability, and parameter inference attacks. However, their methods rely on exhaustive searches. To deal with the involved complexity, the authors restrict their searches to subsamples of the synthetic data. Our method, by representing data in a lower-dimensional euclidean space, drastically reduces the computational burden of finding vulnerable records. Giomi et al.'s work is unique in directly modeling legally outlined risks under a no-box threat model (i.e., the attacker has access to synthetic data, but not the underlying generative model). This makes the approach particularly in line with policy and realistic attack scenarios (cf. Section~\ref{sec:priv}).

Besides this direct use in no-box attacks, distances can be used in attacks with more advanced threat models. In~\cite{meeus2023achilles}, the authors introduce a vulnerability score based on dissimilarity with nearest neighbors. Subsequently, they conduct MIAs with vulnerable synthetic records as targets. Similar approaches exist in which vulnerable records are detected through overfitting detection~\cite{vanbreugel2023membership, MIAnonOutlier}. Our method of representing information in low-dimensional euclidean spaces can increase the efficiency of such methods.

\section{Background}

\subsection{Synthetic Data and Privacy}\label{sec:priv}

The General Data Protection Regulation (GDPR)~\cite{gdpr, ec2016general}, enforced in the Europe Union, is a seminal document in data privacy. The GDPR defines the concept of anonymous data as 
\begin{center}
    \textit{"information that does not relate to an identified or identifiable natural person or to personal data rendered anonymous in such a way that the data subject is not or no longer identifiable."}
\end{center}
The possibility of re-identification, or the process of transforming anonymised data back into personal data, is assessed based on its likelihood within a given dataset~\cite{Stalla}. The Article 29 Working Party (WP29) identifies three key reidentification risks, or \emph{attack types}~\cite{wp29anon}, namely: 1) singling out, or isolating a specific data subject's record(s); 2) linkability, attacking data subjects in one dataset by linking their records to records in another available dataset; and 3) inference, inferring information about some attributes of a data subject. 

\subsection{Notation}
We denote a real dataset by $D$ and a synthetic dataset by $\hat{D}$. By $A(D)$, we denote the set of attributes of $D$, so that every row $d\in D$ ia an $|A(D)|$-tuple with one value for each attribute $a\in A(D)$. We denote by $\mathcal{D}$ the space of all possible rows with attribute set $A(D)$.

\subsection{Distance-Based Privacy Indicators}

Similarity-based privacy metrics are often invoked to measure the degree of privacy protection of synthetic datasets (see, e.g.~\cite{boudewijn2023privacy, OtherSurvey}). The underlying notion is that synthetic records that are too similar to specific real records put them at risk. Synthetic records highly similar to specific real records can for instance lead to identity disclosure (as in the singling out attack in~\cite{Anonymeter}), or the inference of a sensitive attribute of a real individual (e.g. the parameter inference attacks in~\cite{Anonymeter, Goncalves, choi2018generating}). 

The most common distance-based privacy indicator is the \textbf{Distance to closest record} (DCR). Let~$\texttt{Dist}:\mathcal{D}\times\mathcal{D}\rightarrow \mathbb{R}$ be a distance metric. Then the DCR compared two distances: the \textit{Synthetic to Real distance} (SRD) and \textit{Real to Real distance} (RRD). For a given synthetic record $\hat{d}\in\hat{D}$, the SRD is the distance to its closest real record. This is formalized in equation~\eqref{eq:SRD}.

\begin{equation}\label{eq:SRD}
    \text{SRD}(\hat{d}) := \min_{d \in D} \text{Dist}(\hat{d}, d) \quad \forall\hat{d} \in \hat{D}
\end{equation}

The RRD is typically computed with a \emph{holdout set}. Thus, the real dataset $D$ is partitioned into two sets $D_1, D_2$ with $D_1\cap D_2=\varnothing$. For a given record $d_1\in D_1$, the RRD is then the distance to its closest record in $D_2$. This is formalized in equation~\eqref{eq:RRD}.

\begin{equation}\label{eq:RRD}
    \text{RRD}(d_1) := \min_{d_2 \in D_2} \text{Dist}(d_1, d_2) \quad \forall d_1 \in D_1
\end{equation}

Suppose that for some synthetic record~$\hat{d}$, we have $\texttt{SRD}(\hat{d})<\texttt{RRD}(d^*)$, for $d^*=\texttt{arg min}_{d \in D} \texttt{Dist}(\hat{d}, d)$. Then $\hat{d}$ may leak information about real record $d^*$, as it is more similar to $d^*$ than any real record in the holdout set. DCR-based metrics quantify the risks thus incurred by statistically comparing the SRD and RRD distributions. While methods based on both descriptive and inferential statistics have been proposed, the former is much more common~\cite{boudewijn2023privacy}.

In particular, DCR metrics often compare quantiles of the SRD and RRD~\cite{boudewijn2023privacy, Ebert, NeurIPS, PhD_Dani, Platzer}. This provides a metric that measures how much more common small SRD values are to small RRD values. In our experiments, we use the euclidean distance to compute the SRD and RRD on preprocessed (numeric and scaled) data. Suppose the real dataset $D$ is divided into $D_1$ and $D_2$ as above, with $D_2$ functioning as the holdout set. We then use quantiles to compute a DCR metric for a given synthetic dataset $\hat{D}$ as in equation~\eqref{eq:DCR}.

\begin{equation}\label{eq:DCR}
    \texttt{DCR}(\hat{D}, D) := \frac{\left|\left\{\hat{d}\in\hat{D}:\texttt{SRD}(\hat{d})<\texttt{RRD}_\alpha\right\}\right|}{\frac{\alpha}{100}\cdot|D_1|}
\end{equation}

where $\alpha$ is a percentage (we use 2\% throughout) and $\texttt{RRD}_\alpha$ is the $\alpha^{th}$ percentile of the RRD distribution. Equation~\eqref{eq:DCR} is the ratio of SRD values below the $\alpha^{th}$ RRD percentile and the number of RRD values that percentile (naturally $\alpha\%$ of $D_1$). 

To ensure that the measure yields a value of zero when no information is leaked and a value of one when the entire data is leaked, we normalize the DCR between its best case (no SRDs are below the percentile) and its worst case (all SRDs are below the percentile), yielding the privacy score in equation~\eqref{eq:PS}. 

\begin{equation}\label{eq:PS}
    \texttt{Privacy Score}(\hat{D}, D)= \frac{\frac{\alpha}{100}(DCR(\hat{D}, D)-1)}{1 - \frac{\alpha}{100}}
\end{equation}
A privacy score of zero indicates no perceived risks, and a value of one indicates that all synthetic records pose risk to real records. It is important to note that in some cases, the metric can yield negative values. For example, if a perfectly generated synthetic data is an independent portion of the data. Adopting this approach then results in an estimator with an expected value of zero, yet with a non-null variance, which may manifest as small negative values. 

\subsection{Empirical Synthetic Data Privacy Assessment}

The degree of privacy protection in synthetic datasets can empirically be assessed by conducting deliberate attacks. In practice, synthetic data attacks often leverage techniques from adversarial machine learning, including MIAs and shadow modeling~\cite{meeus2023achilles, jordon2022synthetic, TAPAS, membership}. These approaches typically require considerable auxiliary knowledge in addition to the synthetic dataset. 

We follow the approach to empirical assessment proposed by Giomi et al.~\cite{Anonymeter}. In their method, an attack is a guess about a record in a real dataset. For singling out, attacks are queries based on \emph{predicates}, i.e. ``\emph{There is exactly one record in the dataset with predicate $X$}''. Available synthetic data can help inform predicates and thereby guesses. For instance, attribute value combinations $X$ in the singling out guess can be based on attribute value combinations that are unique in the synthetic dataset.

Giomi et al.~\cite{Anonymeter} further propose a statistical assessment method for attack effectiveness. Let ${\bf g}:=\left\{g_1,g_2,...,g_{N_A}\right\}$ be a set of $N_A\in\mathbb{N}$ guesses and ${\bf o}=\left\{o_1,o_2,...,o_{N_A}\right\}$ their respective outcomes, where $o_i$, $i=1,2,...,N_A$ is defined through equation~\eqref{eq:o}.
\begin{equation}\label{eq:o}
o_i:=\begin{cases}
    1, &\text{if guess $g_i$ was correct}\\
    0, &\text{otherwise}
\end{cases}
\end{equation}
Guess effectiveness can then be interpreted as a Bernoulli trial, with some success probability $\hat{r}$. By measuring number of successful guesses, estimators $r, \delta_r$ can be computed such that $\hat{r}\in r\pm\delta_r$ at confidence level~$\alpha$ via the Wilson Score Interval, using equation~\eqref{eq:estimate}.
\begin{equation}\label{eq:estimate}
\begin{array}{ll}
r&:=\frac{N_S+z_{\alpha}^2/2}{N_A+z_{\alpha}^2}\\
&\\
\delta_r&:=\frac{z_{\alpha}}{N_A+z_{\alpha}^2}\sqrt{\frac{N_S(N_A-N_S)}{N_A} + \frac{z_{\alpha}^2}{4}}
\end{array}
\end{equation}
where $N_S$ is the number of correct guesses.

To contextualize attack effectiveness, Giomi et al.~\cite{Anonymeter} proceed as follows. Prior to training, divide the real dataset into a training set and a control set. Use the training set to construct the synthetic data. After training, attack the training dataset through privacy attacks conducted by building predicates on the synthetic data. Let~$r_{\texttt{train}}$ denote the estimator of the success rate of attacks informed by the synthetic data. Next, conduct attacks on the control set by building predicates on the synthetic data. Let~$r_{\texttt{control}}$ denote the success rate of these latter attacks. Risk is computed using equation~\eqref{eq:baseline}.

\begin{equation}\label{eq:baseline}
    R := \frac{r_{\texttt{train}} - r_{\texttt{control}}}{1 - r_{\texttt{control}}}
\end{equation}

The difference between~$r_{\texttt{train}}$ and~$r_{\texttt{conrol}}$ indicates how much of the success of the attacks is due to the generator memorizing specific attributes from the training data. The denominator is a normalization factor, with $1$ being the largest possible value attained by~$r_{\texttt{train}}$.

\subsection{Contrastive Learning}

Contrastive Learning is a machine learning technique that aims to learn meaningful representations of data in a self-supervised manner. Unlike traditional methods, it doesn't rely on labeled information during the learning process. Instead, it endeavors to bring similar samples closer together in the representation space while pushing dissimilar samples farther apart. In the context of tabular data, a method for acquiring low-level representations is detailed in the work by Shenkar et al.~\cite{shenkar2021anomaly}. Their method divides rows of a dataset into non-overlapping subrows. When properly trained, their mappings locate both subrecords of a given record closely together in the embedding space. 

\section{Methods}

\subsection{Constrastive learning approach}

\begin{figure*}[ht]
       \centering
        \includegraphics[height=0.25\textheight]{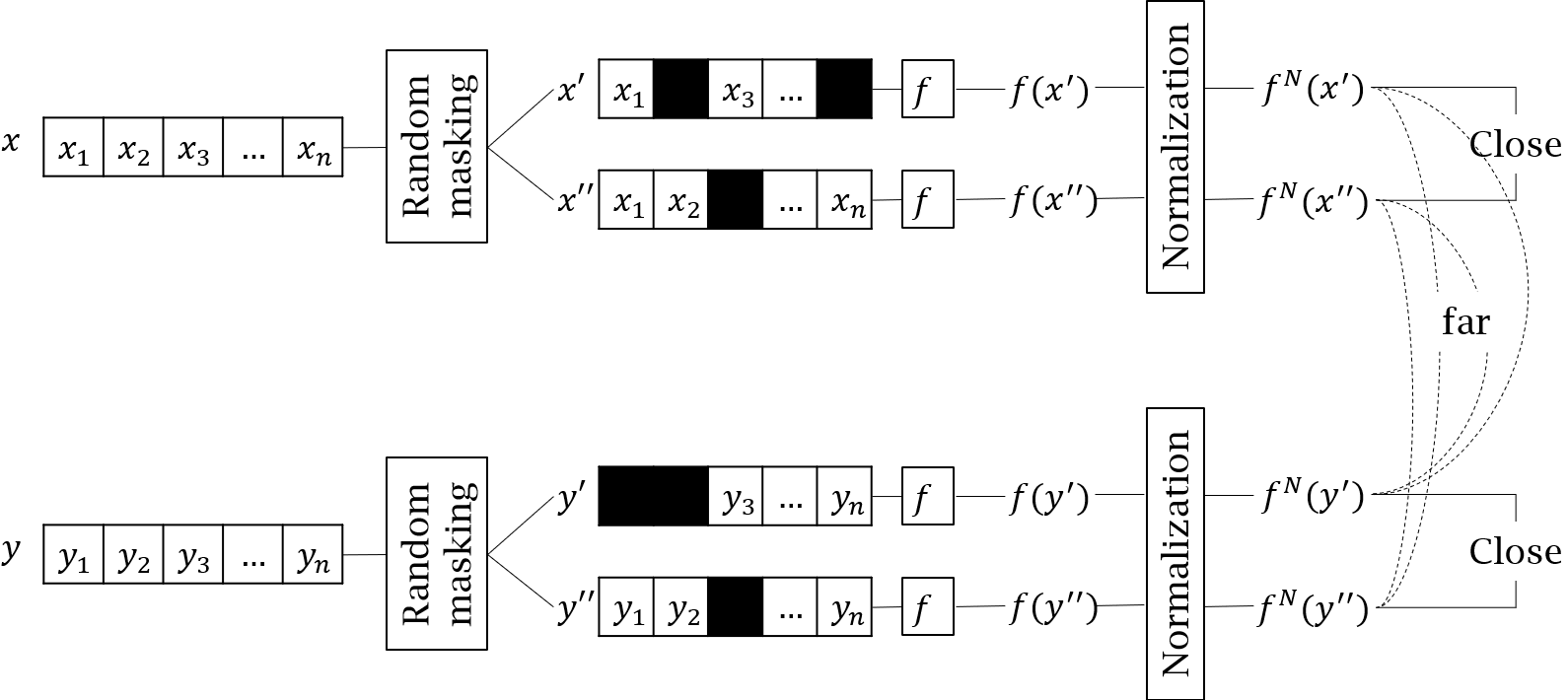}
        \caption{Contrastive learning approach}
        \label{fig:contrast}
    \end{figure*}

Our contrastive learning method applies a random masking function to each record $d\in D$. Subsequently, it is tasked to recognize the similarity between two distinct maskings of the same record. Let $x,y\in D$, with $x=(x_1,x_2,...,x_n)$. Let $x'$ be a masking of $x$, i.e. a tuple $(x_1',x_2',...,x_n')$, with 
$$
x_i'=\begin{cases}
    x_i, &\text{if column $i$ is not masked}\\
    \emptyset, &\text{otherwise}
\end{cases}
$$
for $\emptyset$ a masked (missing) value. Suppose $x', x''$ are two possibly distinct maskings of $x$ and $y', y''$ are two possibly distinct maskings of $y$. We then aim to learn a neural network $f$ that maps rows of $D$ to the embedding space, such that for all $x,y\in D$ and for a given similarity metric $\texttt{sim}$, we have that $\texttt{sim}(f^N(x'),f^N(x''))$ and $\texttt{sim}(f^N(y'),f^N(y''))$ are close to one, but $\texttt{sim}(f^N(x^a),f^N(y^b)),$ $a,b\in\left\{',''\right\}$ are close to zero, where $f^N(z)$ denotes $f(z)$ after normalization. This is illustrated in Figure~\ref{fig:contrast}.

We use a stochastic masking function, choosing at random anywhere from one to $n-1$ columns to mask. We implement $f$ as a simple neural network. The dimension $m$ of the embedding layer is determined based on the number of attributes in the dataset. To mitigate overfitting, a dropout layer is applied with a probability of 0.1 after a normalization layer. The GELU activation function is utilized throughout the network. Categorical variables undergo an initial embedding layer before being processed through the aforementioned network. We use cosine similarity as the similarity metric in the normalized embedding space, using cross entropy loss to train $f$. For each $x,y\in D$, we compare $f(x')$ to $f(x'')$, and to one (rather than both) of $f(y')$, $f(y'')$ to balance similar and dissimilar comparisons.

\subsection{Evaluation}

To assess the effectiveness of our method, we use vulnerable records as targets in deliberate attacks. For each given identified outlier, we construct a predicate as follows. For categorical attributes, we take the category of the outlier. For numeric attributes, we divide their range into finitely many bins. The predicate then takes the bin (interval) in which the outlier value occurs as the attribute's range. Multivariate predicates are constructed by concatenating univariate predicates. Predicates that single out unique synthetic data records are used to mount attacks on the train and control sets. Following~\cite{Anonymeter}, we evaluate attack effectiveness through equation~\eqref{eq:baseline}. We conduct two types of experiments: 1) ``leaky evaluation'': experiments using \emph{leaky datasets}; 2) ``deep-learning evaluation'': experiments using synthetic data generated by deep-learning models. For the attack-based metrics, hyperparameters are consistent with those in~\cite{Anonymeter} throughout.

\textbf{Leaky evaluation.} Experiments using leaky datasets proceed as follows. The real dataset $D$ is divided into three sets: training, control, and release. The ``synthetic'' dataset $\hat{D}$ is built using a fraction of records from the training and the remaining ones from the release set. The fraction is denoted by $f_l$. When $f_l$ is set to $0$, the whole release set is used as the synthetic dataset. When $f_l = 1$ the synthetic set is a copy of the training one. By exploring different levels of leak fraction, this test shows how the singling out risk increases linearly with the leak fraction. 

\begin{figure}[ht]
       \centering
        \includegraphics[height=0.2\textheight]{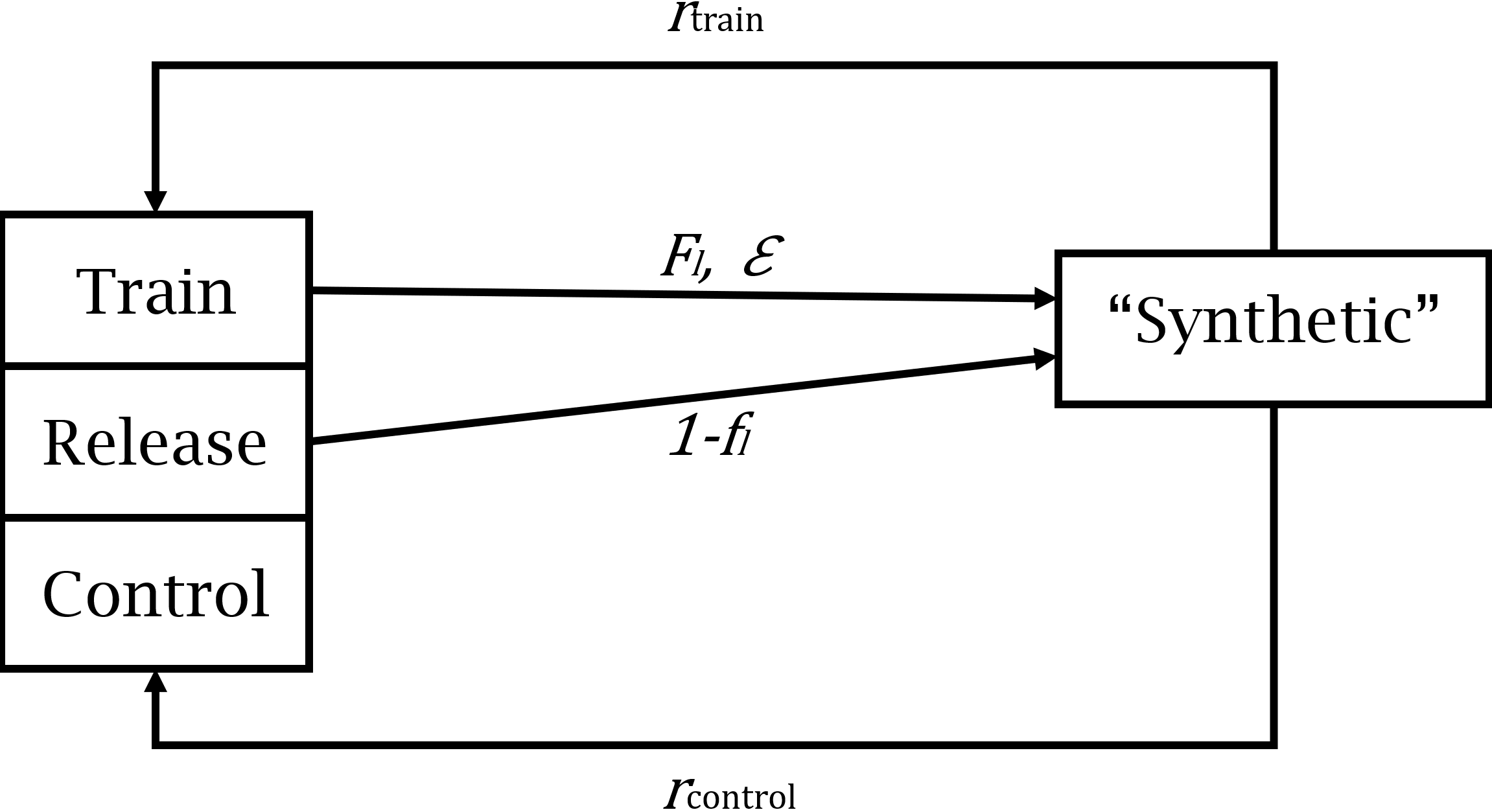}
        \caption{Leaky evaluation, where $\varepsilon$ denotes noise addition}
        \label{fig:linear}
    \end{figure}

To assess the robustness of the results, we inject noise to the training records in $D'$. For numerical floating point attributes, the noise is sampled from a Gaussian distribution $\mathcal{N}(0,\sigma)$, for user-defined standard deviation~$\sigma$. For numerical integer data types, noise is sampled from a Poisson distribution with parameter $\lambda$. For categorical attributes, the category is changed at random with probability $p$. This is illustrated in Figure~\ref{fig:linear}. 

\textbf{Deep-learning evaluation.} Experiments conducted with synthetic datasets generated by deep learning models provide more insight into the metrics' performance in practice. Three models are used to obtain the required synthetic data: CTGAN~\cite{xu2019modeling}; DPCTGAN~\cite{fang2022dp}; and RealTabFormer~\cite{solatorio2023realtabformer}. In synthetic data generated by deep-learning models, overfitting is a main cause of privacy risks. To assess the metrics' accuracy in identifying such risks, we conduct experiments with synthetic data stemming from deliberately overfit models. 

To do so, we first train the RealTabFormer optimally and store the minimal value of its validation loss function, denoted by $L^*$. Subsequently, we let $f_o\in\left[1, 2\right]$ denote the \emph{overfitting ratio}. We then train the RealTabFormer again on the same data, but prematurely terminate the training once its validation loss function $L$ is such that $\frac{L}{L^*}=f_o$. When $f_o=1$, we have $L=L^*$, so no deliberate overfitting took place. When $f_o=2$, the training is terminated when the loss function is twice the previously discovered minimum $L^*$.

\section{Experiments}

\subsection{Datasets}

This methodology has been tested on three different datasets, namely: Adult \cite{misc_adult_2}, the Texas Inpatient Public Use Data File (``Texas'')~\cite{misc_texas}, and the 1940 Census full enumeration from IPUMS USA (``Census'')~\cite{misc_census}. The adult dataset has 48842 rows and fifteen attributes, of which six are numerical (float), and the remaining nine are categorical. The Texas and Census datasets are considerably larger, at respectively 193 and 97 attributes. To benchmark our results against those of Giomi et al.~\cite{Anonymeter}, we limit our experiments to the same subsets  of 28 and 37 attributes for these sets, respectively. Likewise, for Texas and Census, we take randomly selected, mutually exclusive subsets of cardinality $20000$ and $25000$, respectively, as training and control sets.

\subsection{Experimental Set-Up}

The implemented contrastive learning networks all comprise three hidden layers, each consisting of 1024 neurons. For datasets with a greater number of columns, a higher embedding dimension $m$ is chosen. In our specific cases, we select dimensions of 10 for the "Adults" dataset, 20 for the "Texas" dataset, and 30 for the "Census" dataset (see Appendix~\ref{app:hyper}). During training, 300 epochs are executed with a batch size of 1024. Early stopping regularization is implemented, with a monitoring window of 20 epochs and a patience of 10. The learning rate is set to $10^{-3}$.

For singling out, we conduct univariate and multivariate experiments with the number of attributes varying between $3$ and $12$, each consisting of at most\footnote{The number of guesses is dependent on how many predicates single out unique synthetic records. This may be fewer than 2000 if no more vulnerable records can be detected.} 2000 attacks ($N_A=2000$). The final reported risk is the highest risk $R$. For linkability and inference, we use the parameters from~\cite{Anonymeter}. For the leaky evaluation, we additionally conducted experiments for all configurations of noise injection parameters $\sigma, \lambda, p\in\left\{0, 0.05\right\}$ and for leak fraction $f_l$ varying between zero and one. We test the contrastive learning method with Giomi et al.'s~\cite{Anonymeter} method as a baseline. Bootstrapping with $n=1000$ is used to measure the variance of the methods.

For the CTGAN and the DPCTGAN, we used the same hyperparameters as in~\cite{Anonymeter}. For the REalTabFormer, we used default learning parameters for the model with no overfit. For the overfitted experiment we disabled the earlystopping and other regularizations (the dropouts in the network). We then ran the framework for 100 epochs and chose the models with the desired fraction of overfitting.

\subsection{Results}

\begin{figure*}[h]
       \centering
        \includegraphics[height=0.8332\textheight]{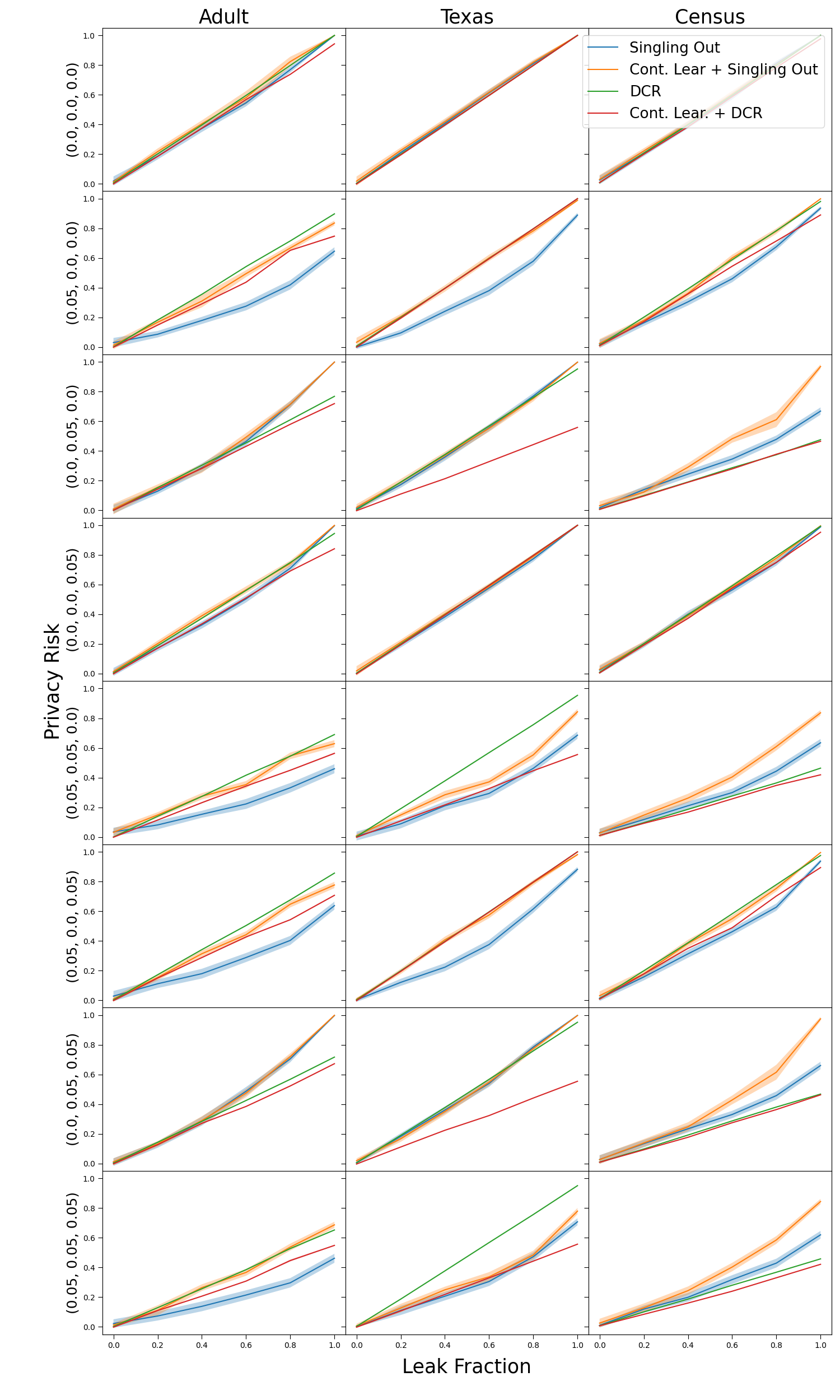}
        \caption{Leaky evaluation results with 95\% confidence interval error bars, by paramterization and dataset. The tuples in the labels of the $y$-axes indicate parameter values of $(\sigma,\lambda,p)$.}
        \label{fig:results}
\end{figure*}

\begin{figure*}[h]
    \centering
    \includegraphics[scale=0.45]{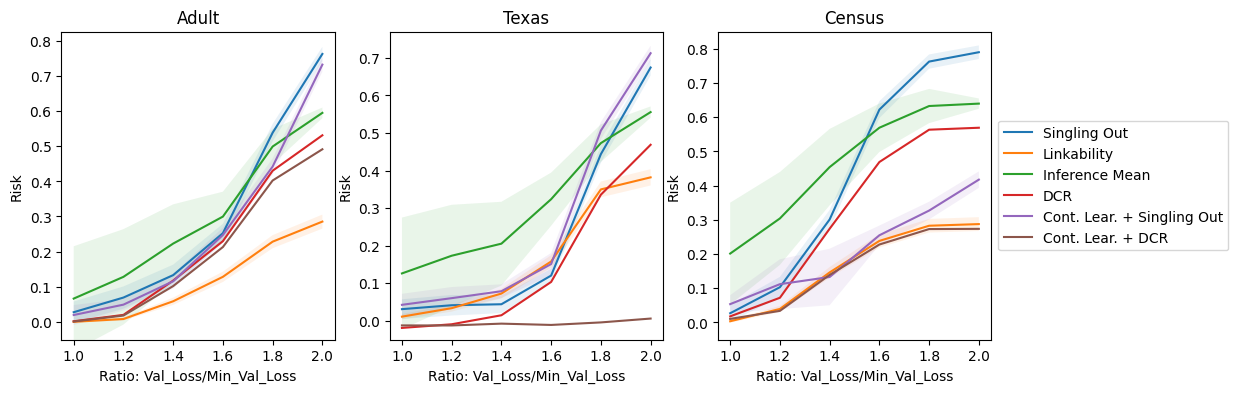}
    \caption{DCR and Anonymeter result with REalTabFormer synthetization. Confidence intervals for DCR are computed via bootstrapping with $n=1000$.}
    \label{fig:Anonymeter_vs_knn_REalTabFormr}
\end{figure*}

For the attack-based metrics, all the reported results refer to the maximum risk $R$ among the ones calculated in the corresponding setting to provide the most stringent risk estimation. Results for the leaky evaluation are provided in Figure~\ref{fig:results}. Regarding the deep-learning evaluation, Table~\ref{tab:res} contains the results of the methods for datasets generated by non-overfit models. Corresponding computation times of the all metrics are provided in Table~\ref{tab:time} (Further details on computational resources and scalability in Appendices~\ref{appendix:comp-res} and~\ref{appendix:scalability}). Figure~\ref{fig:Anonymeter_vs_knn_REalTabFormr} shows the results of the deep-learning evaluation with varying degrees of deliberate overfitting.

    \begin{table*}[h]
    \centering
    \begin{tabular}{llllll}
        \toprule
         Dataset                  & Method   & SO~\cite{Anonymeter} & SO + CL & DCR & DCR + CL\\
         \midrule
           \multirow{3}{*}{Adult} & CTGAN &  0.12431 $\pm$ 0.018 & 0.08193 $\pm$ 0.024 & -0.00843 $\pm$ 3.9 $\times 10^{-4}$ & -0.01048 $\pm$ 4.4 $\times 10^{-4}$ \\
                                  & DPCTGAN &  0.11184 $\pm$ 0.017  &  0.09916 $\pm$ 0.015   & -0.01993 $\pm$ 2.2 $\times 10^{-5}$ & -0.01996 $\pm$ 2.3 $\times 10^{-5}$ \\   
                                  & REalTabFormer & 0.02783 $\pm$ 0.031 & 0.01984 $\pm$ 0.028 & 0.00467 $\pm$ 0.001 & -0.01815 $\pm$ 1.1 $\times 10^{-4}$\\ \midrule

           \multirow{2}{*}{Texas} & CTGAN   &  0.01537 $\pm$ 0.015                                &    0.01115 $\pm$ 0.009  & -0.02020 $\pm$ 2.6 $\times 10^{-5}$    &  -0.01061 $\pm$ 6.0 $\times 10^{-4}$  \\
                                  & DPCTGAN &  0.00816 $\pm$ 0.009                                &   0.00611 $\pm$ 0.009   & -0.02040 $\pm$ 0.0   &  -0.01245 $\pm$ 2.7 $\times 10^{-4}$    \\
                                  & REalTabFormer & 0.03103 $\pm$ 0.026 & 0.04340 $\pm$ 0.029 & -0.01893 $\pm$ 9.6 $\times 10^{-5}$ & -0.01194 $\pm$ 1.9 $\times 10^{-4}$\\ \midrule

           \multirow{2}{*}{Census} & CTGAN  &  0.01340 $\pm$ 0.021                           &  0.00783 $\pm$ 0.005    & -0.01961 $\pm$ 4.3 $\times 10^{-5}$     & -0.01943 $\pm$ 1.0 $\times 10^{-4}$  \\
                                  & DPCTGAN & 0.01077 $\pm$ 0.009                               &    0.00871 $\pm$ 0.005   & -0.02040 $\pm$ 0.0 & -0.01999 $\pm$ 3.6 $\times 10^{-5}$ \\ 
                                  & REalTabFormer & 0.02695 $\pm$ 0.024 & 0.04650 $\pm$ 0.027 & 0.01882 $\pm$ 4.88 $\times 10^{-4}$ & 0.01008 $\pm$ 5.1 $\times 10^{-4}$\\ 
        \bottomrule
    \end{tabular}
    \caption{Measured risks for deep-learning-based synthetic datasets. SO: Singling out attack from~\cite{Anonymeter}; DCR: distance to closest record; CL: contrastive learning. For the Texas and Census datasets, DPCTGAN, DCR synthetics data has no record with SRD below the $RRD_{\alpha}$, so there no variability in the risk measure during the bootstrapping procedure.}
    \label{tab:res}
\end{table*}

\begin{table*}[h]
    \centering
    \begin{tabular}{lllllll}
        \toprule
         Dataset   & DCR & DCR + CL & SO & SO + CL & Linkability ~\cite{Anonymeter} & Inference ~\cite{Anonymeter} \\
         \midrule 
           Adult   & 6.91 & 238.16 &  2842.78&863.95 &  18.06 &  163.23 \\             
           \midrule
           Texas   & 134.72 & 460.65 &  376.82 &1176.21 & 103.01 & 1380.70     \\        
           \midrule
           Census  & 83.35 & 364.40 &  2167.73&1645.03 & 99.26 & 1603.49\\           
        \bottomrule
    \end{tabular}
    \caption{Measured time (in seconds) for different attacks. SO: Singling out attack from~\cite{Anonymeter}; DCR: distance to closest record; CL: contrastive learning}
    \label{tab:time}
\end{table*}

\section{Conclusion}

\subsection{Discussion}

In all experiments (leaky evaluation and overfitting evaluation), the DCR methods showcased the same response to deliberate privacy leaks as the attack-based methods. In the leaky evaluation, indicated risk consistently increased linearly with the leak fraction. This behavior was robust against deviations in attribute values. In the deep-learning evaluation, the increase in measured risk with the degree of deliberate overfitting followed a similar pattern for all metrics. The only exception was the DCR evaluated in embedded space for the Texas dataset, which failed to measure significant risks. 

In all experiments, the DCR metrics had considerably smaller standard deviations than the attack-based methods. This indicates that their performance is more consistent between experiments. This was especially true in the deep-learning evaluation, where attack-based metrics' standard deviations were particularly large. Such metrics can therefore not guarantee that all present risks are detected with the same certainty as the DCR metrics. Results further underscore the efficiency of the DCR, computed in $2\cdot10^{-2}$ the amount of time as singling out attacks on average ($2\cdot 10^{-1}$ when using contrastive learning-based embeddings).

The leaky evaluation showed an increased sensitivity and robustness of singling out attacks to parameter changes when embeddings were invoked. This was particularly notable when values of integer attributes were altered and for large leak fractions. The effect was strongest in the experiments with the Census dataset, but also present in the other experiments. Interestingly, the experiments with the Texas dataset showed the opposite effect for the DCR. This robustness to data fluctuations is important, as it may indicate how well equipped the privacy metrics are at dealing with data subject to drift. Interestingly, the impact of embeddings was less pronounced in the deep-learning evaluation.

Overall, we find that both the DCR and the attack-based metrics increase with leakage and overfitting in a similar fashion. The impacts of both the metrics and contrastive learning-based embeddings vary with the involved dataset and manner of modeling privacy breaches (leaks or overfitting). The DCR was the most efficient metric. Data types and the sensitivity of attributes may have a strong impact on the results. This is evidenced by the strong discrepancies that occurred in the leaky evaluation, when integer values were distorted. The DCR proved to measure risks in a manner consistent with the attack-based metrics modeled after the WP29 definition of privacy. Furthermore, their results were more consistent and their computation was much more efficient. Use of contrastive learning-based embeddings can improve the performance of distance-based privacy metrics, making them more robust to attribute value fluctuations. 

\subsection{Future Research}

Future research should identify the conditions under which embeddings improve the performance of privacy metrics. In particular, the impact of data types and attribute value changes on measured risk should be more clearly understood. The difference in performance between leaky and deep-learning evaluations should also be studied, as should potential causes of failure for the DCR in embedded spaces. This latter help clarify why the performance of this approach deviated in one of our deep-learning evaluation experiments. 

Future research should further combine embedding methods with other outlier detection methods (see, e.g.,~\cite{Chandola, Pang, Ruff_2021, isolation}) to assess the role of representation in vulnerable record discovery, perticularly in membership inference attacks targeting vulnerable records. This could lower the computational burden of such attacks by limiting the scope to relevant potential targets. It could also extend the scope of attacks to include targets not directly obvious without using embeddings.

\bibliographystyle{IEEEtran}
\bibliography{IEEEabrv, mybibliography}

%



\appendices
\section{Computational Resources and Reproducibility}\label{appendix:comp-res}
In our experiments, we utilized computational resources comprising: 

\begin{itemize}
    \item CPU: AMD Ryzen Threadripper 2950X 16-Core Processor
    \item GPU: NVIDIA RTX A5000 
    \item RAM: 128 GiB
\end{itemize}
These hardware configurations provided the necessary computational power and memory capacity to execute our experiments efficiently. Additionally, smaller-scale experiments were conducted using a MacBook Air equipped with an M2 chip and 16 GB of RAM. This setup ensured that the model could operate effectively even with limited computational resources, particularly when the dataset under analysis could be accommodated within the available memory.
To facilitate reproducibility, the source code used in this study is available on GitHub at \url{https://github.com/aindo-com/privacy-eval-paper}.

\section{Scalability}\label{appendix:scalability}
To assess the scalability of our method, we conducted a study using the Census dataset. The provided table displays the experiment runtimes relative to the dataset's row count. Synthetic data was generated using the REalTabFormer framework \cite{solatorio2023realtabformer}.
\begin{center}
    \begin{tabular}{|c|c|c|c|c|}
        \hline
         \#Rows & 25000 & 50000 & 75000 & 100000  \\
         \hline
         SO \cite{Anonymeter} &  12906.35 & 17124.45 & 21748.37 & 26377.15 \\
         \hline
         SO + CL & 1467.05 & 13238.44 & 17256.22 & 23583.04 \\
         \hline
    \end{tabular}
\end{center}

This demonstrates the superior scalability of our method compared to the Anonymeter framework on small sized datasets. Further investigations into scalability on larger datasets are necessary to evaluate whether our proposed solution outperforms current state-of-the-art approaches.

\section{Hyperparameters Analysis} \label{app:hyper}

To choose the best embedding dimension, we explored how different dimensions impact the performance of our contrastive learning model, particularly focusing on accuracy, loss, and an additional risk measure obtained after training. We tested dimensions ranging from $3$ to $N-1$ where $N$ is the total number of features in our dataset. The following table shows the results for the Adult dataset.

\begin{center}
    \begin{tabular}{|c|c|c|}
        \hline
         Embedding Dim & Risk & CI \\
         \hline
         3 & 0.0089 & 0.0134  \\
         \hline
         4 & 0.0124 & 0.0121 \\
         \hline
         5 & 0.0168 & 0.0281 \\
         \hline
         6 & 0.0074 & 0.0303 \\
         \hline
         7 & 0.0098 & 0.0113 \\
         \hline
         8 & 0.0152 & 0.0289 \\
         \hline
         9 & 0.0087 & 0.0307 \\
         \hline
         10 & 0.0243 & 0.0272 \\
         \hline
         11 & 0.0078 & 0.0277 \\
         \hline
         12 & 0.0167 & 0.0303 \\
         \hline
         13 & 0.0123 & 0.0180 \\
         \hline
         14 & 0.0066 & 0.0281 \\
         \hline
    \end{tabular}
\end{center}
The analysis of the results reveals that the maximum computed risk is observed when the embedding dimension is set to 10. However, it is important to highlight that, when accounting for the confidence intervals, there is no statistical evidence to conclude that the risk associated with one embedding dimension is definitively higher or lower than that of another. This suggests that while the maximum risk occurs at dimension 10, other dimensions may yield similar risk levels, and any differences may be within the margin of error.

\begin{figure}[H]
    \centering
    \includegraphics[width=1\linewidth]{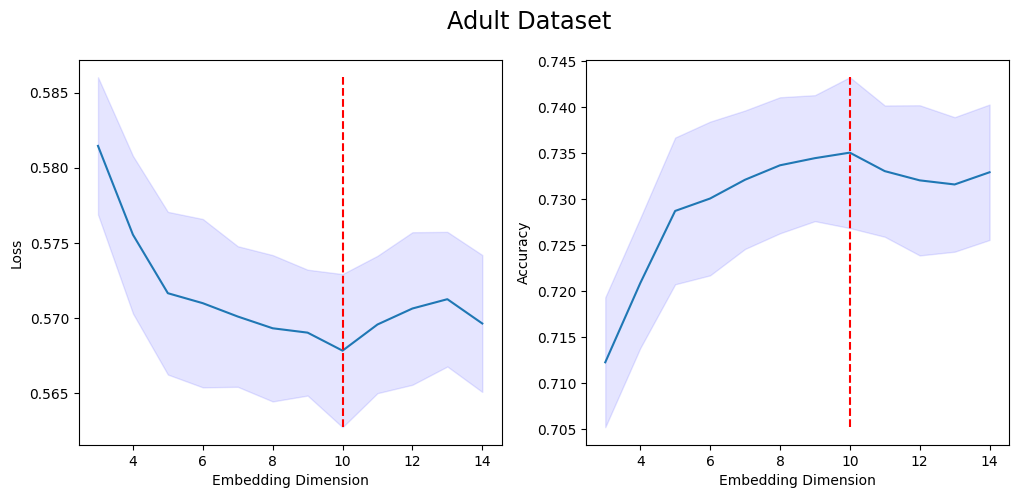}
    \caption{Validation loss and accuracy for different embedding dimensions}
    \label{fig:vloss-acc}
\end{figure}

Figure ~\ref{fig:vloss-acc} presents a detailed view of how the accuracy and validation loss evolve as the embedding dimension increases. Along with these trends, the 95\% confidence intervals are plotted to emphasize that, statistically, no particular embedding dimension outperforms others in a significant way, apart from the early dimensions. Specifically, the initial dimensions tend to show higher validation losses and lower accuracy values, indicating that they may not be optimal for the task at hand.

Furthermore, the red dashed line in the figure represents the best embedding dimension based solely on the point estimate, which does not take into account the variability captured by the confidence intervals. This line highlights the embedding dimension that produces the lowest risk according to the point estimate, but, as the confidence intervals show, other dimensions may be just as effective in practice. Thus, the overall conclusion is that, despite the apparent superiority of the embedding dimension at 10, there is no strong statistical justification for claiming that one dimension is conclusively better than another within the considered range.

We used the same approach to pick the best embedding dimension for both the Texas and Census datasets. As in the previous case, there wasn’t enough statistical evidence to clearly favor one specific dimension. So, in the end, we went with the dimension that had the best point estimate, acknowledging that while this choice looked optimal, it’s not statistically more reliable than the nearby dimensions.

\end{document}